\documentclass[runningheads]{llncs}

\usepackage[utf8]{inputenc} % allow utf-8 input

\usepackage[]{algorithm2e}
\usepackage{amsmath, amssymb}
\usepackage{bbm}
\usepackage{todonotes}
\usepackage[hidelinks]{hyperref}

\usepackage[caption=false]{subfig}

\usepackage{xspace}
\newcommand{\C}[1]{\mathcal{#1}}

\usepackage{mathabx}

\def\bf{\bfseries}

\usepackage{xargs}
\usepackage{mathtools}
\newcommandx\val[3][2,3,usedefault]{{#1_{#2}^{#3}}}

\DeclareMathOperator*{\argmin}{arg\,min}
\DeclareMathOperator*{\argmax}{arg\,max}

\usepackage{biblatex}
\begin{document}

\title{Sample Noise Impact on Active Learning}

\author{Alexandre Abraham\inst{1}\orcidID{0000-0003-3693-0560}\\ \and
 Léo Dreyfus-Schmidt\inst{1}\orcidID{0000-0001-8271-1217}}

\authorrunning{A. Abraham et al.}

\institute{Dataiku, Paris, France\\
\email{\{alexandre.abraham,leo.dreyfus-schmidt\}@dataiku.com}}

\maketitle

\begin{abstract}
  This work explores the effect of noisy sample selection in active learning strategies.
  We show on both synthetic problems and real-life use-cases that knowledge of the sample noise can significantly improve the performance of active learning strategies. Building on prior work, we propose a robust sampler, \textit{Incremental Weighted K-Means} that brings significant improvement on the synthetic tasks but only a marginal uplift on real-life ones.
  We hope that the questions raised in this paper are of interest to the community and could
  open new paths for active learning research.

\end{abstract}

\section{Introduction}
When training machine learning models, data quality is undoubtedly the most fundamental
requirement.  A recent study \cite{northcutt2021pervasive} has shown that pervasive
errors in the test set of famous datasets could lead to selecting a suboptimal model.
In active learning, where a small number of samples are selected to be 
labeled by an oracle, it becomes paramount as selecting samples of poor quality 
may worsen the model's performance.

Sample diversity in the training set is also essential and has been the main focus of
recent active learning strategies. 
Performance improvements come from new ways of combining uncertainty and diversity in
a single framework. BatchBALD \cite{kirsch2019batchbald} adds diversity by minimizing the joint mutual information between 
batch samples. Core-sets \cite{sener2017active} and \cite{zhdanov2019diverse}
use a clustering approach to scatter the selected samples across the sample space. The method
proposed in \cite{du2015exploring} minimizes the similarity between the samples of the batch
while minimizing the similarity with already labeled samples. The most common explanation 
for the observed performance uplift when enforcing diversity is that a homogeneous set of samples contains
much redundant information while a diverse one informs the model with several
classification patterns.

Enforcing diversity entails selecting samples where uncertainty
is not maximal. Therefore, the selected samples are further away from
the decision boundary and easier to classify. We hypothesize that this side-effect of diversity
contributes to its success. In classification, mislabeled or very ambiguous samples
--~like five that looks like six in MNIST~-- can be detrimental to the model \cite{northcutt2021pervasive}. As the density of such
samples is higher near the classification boundary, we increase the chances of obtaining meaningful samples by selecting samples further away.

This paper proposes a metric to evaluate the quantity of such noisy samples in a dataset, and we design a query strategy
to avoid them. We first validate our approach by showing the existence of these samples on a synthetic example and observe that diversity-based methods are less
likely to select those. We show that our results obtained on synthetic data do not generalize well to real tasks,
propose an explanation and ideas to mitigate the problem.

\section{Sample-noise robust strategies}

In the following, $\C{D}$ designates a dataset and $h$ a probabilistic classifier. A subscript indicates the nature of datasets: $L$ stands for labeled samples, $U$ unlabeled, $T$  test, and $B$ designates a batch of samples. Iterations are indicated with a superscript when pertinent.

\subsection{The pervasiveness of sample noise}

In his seminal work on active learning, Settles \cite{settles2009active} defines
the most valuable samples at iteration $i$ as the one with the lowest maximum predicted probability among classes:

$$\text{lowest\_confidence(x)} = 1 - h^i_{1}(x)$$

With $h^i_k$ being the $k$-th probability predicted by the classifier learned at iteration $i$ in descending order, so that $h^i_{1}$ is the maximum predicted probability at iteration $i$. This definition assumes that each sample can reach a predicted probability of $1$. The difference between $1$ and the predicted probability represents the information that the model is expected to gain when the sample gets labeled.

\begin{figure}[htp]
  \centering
  \includegraphics[width=\textwidth]{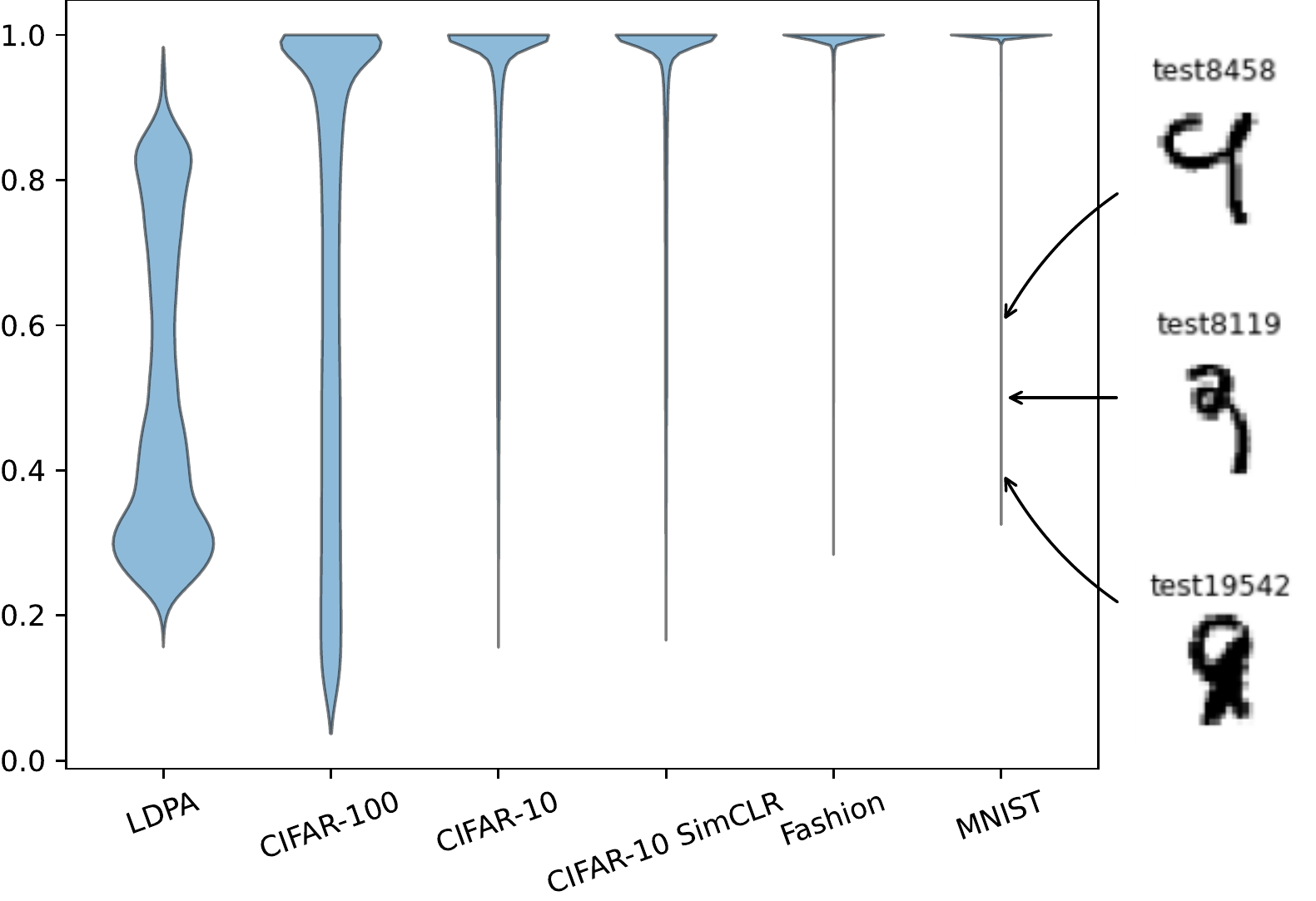}%

  \caption{Distribution of prediction probabilities by a model in a 2-fold setting, and examples of ambiguous samples on the MNIST dataset.}
  \label{fig:ambiguous}
  
\end{figure}

However, classifiers do not always reach a predicted probability of $1$ for all samples. Fig.~\ref{fig:ambiguous} shows the distribution of predicted probabilities on various standard tasks (see details in section~\ref{sec:exp}). If some datasets like LDPA present an almost uniform distribution, MNIST is very polarized towards $1$ while having outliers below $0.5$.\\

We call noisy the samples located at the boundary between two classes, which commonly have a low predicted probability for their class. Noisy samples can be due to signal noise in the data that makes them hard-to-classify, labeling errors, or to a genuine ambiguity such as a four that looks like a nine in MNIST (see Fig. \ref{fig:ambiguous}, right). Noisy samples are a challenge in active learning as they may get overly selected by uncertainty-based methods despite their low quality. At a given iteration of an active learning experiment, noisy samples occur for two reasons. First, those samples may be easy to classify, but our current classifier lacks the knowledge to do so. Labeling this sample could be useful as it would help the model determine if the ideal decision boundary is close or not. This type of uncertainty is called \textit{epistemic} and can be reduced with more samples. However, it may also be that this sample is ambiguous and that an ideal classifier would not do any better. The noise is then due to aleatoric uncertainty that cannot be reduced.

Let us call $h^{\infty}$ this ideal classifier obtained by training the model on all available training data . We use it to define the theoretical \textit{informed lowest confidence} sampler (denoted by \textit{IConfidence}) based on the following score:

$$\iota(x) = h^{\infty}(x) - h^i_{1}(x)$$

We expect this sampler to account for aleatoric uncertainty and therefore focus only on reducing epistemic uncertainty. If $h^{\infty}$ is unknown at experiment time, it can be estimated in a research context where all labels are known. Such an oracle can be useful in active learning research by providing a golden standard of the maximum achievable accuracy in an experiment.

\subsection{Measuring sample noise}

Misclassified samples are a source of sample noise, and \cite{northcutt2021pervasive} proposes to identify them using human annotation. This approach can be considered a golden standard but is hard
to perform because of human labeling costs.

We previously suggested that sample noise could be measured as the maximum probability predicted by a good enough classifier.
In order to extend this measure to a set of samples, we propose to rely on a metric previously introduced in \cite{abraham2020rebuilding}
called \textit{reverse batch accuracy} or RBA for short. RBA measures how easy samples are to classify by training a classifier on the test
set and measuring its accuracy on sample batches. The lower the RBA score, the harder
samples are to classify for the model, so the noisier are the samples.

\subsection{Incremental Weighted K-Means (IWKMeans)}

The goal of batch active learning strategies is to select batches of samples 
$\C{D}_B$ representative of the unlabeled data $\C{D}_B \sim \C{D}_U$. For a given notion of 
similarity $\text{sim}$ between batches, this leads to the following maximization objective:
\begin{equation}
  \label{eq:sim}
\text{argmax}_{\C{D}_B} \text{sim}(\C{D}_B, \C{D}_U)
\end{equation}

In \cite{zhdanov2019diverse}, the similarity is taken as $- \sum_{u \in \C{D}_U} d(\C{D}_B, u)$ with $d$ being the squared distance to the closest
point in the set $d(\C{D}_B, u) = \min_{b \in \C{D}_B} \|b - u\|^2$. This corresponds to the inertia objective of the K-Means clustering.
The authors propose to use it in a two-step procedure called \emph{Weighted K-Means} (WKMeans)
where a set of samples are preselected using margin sampling, and then the final batch is selected by using
K-Means.

The above objective does not consider already labeled data and can lead to suboptimal batches lying in regions of high-density of labeled samples. A natural refinement is to additionally impose that the selected batch differs from already labeled data, \textit{i.e.} to minimise similarity $ \text{sim}(\C{D}_B, \C{D}_L)$:
\[
\text{argmax}_{\C{D}_B} \text{sim}(\C{D}_B, \C{D}_U) \quad\quad \text{subject to argmin}_{\C{D}_B} \text{sim}(\C{D}_B, \C{D}_L)
\]
In the context of K-Means, minimizing this similarity is equivalent to preventing points close to labeled data to \emph{drag} the centroids toward them. This is done by adding the labeled points in the reference set used to compute distances in the K-Means objective that becomes $- \sum_{u \in \C{D}_U} d(\C{D}_B \cup \C{D}_L, u)$. This translates algorithmically by adding cluster centers corresponding to already labeled samples and keeping them fixed during optimization. 
We refer to this approach as \textit{Incremental Weighted K-Means} or IWKMeans for short, and it is described in Alg.~\ref{alg:iwkmeans}. IWKMeans tends to \emph{repel} batch samples from already selected samples, including the noisy ones. A similar approach is proposed in \cite{du2015exploring} where the values
in the matrix of similarity between batch and selected samples are minimized.
 
\begin{algorithm}[hbtp]
  \KwData{$\C{D}_L^0, \C{D}_U^0$}
  \KwResult{$h^{n_{iter}}$}
  \For{$i\leftarrow 1$ \KwTo $n_{iter}$}{
    Margin sampling to pre-select $\beta k$ samples among the unlabeled ones:
      $P^i = \argmax_{\C{D}_U^i} 1 - (h^i_{1}(x) - h^i_{2}(x))$ \\
    Perform K-Means on $P^i$ with $k$ moving and $\C{D}_L^i$ fixed centroids:
      $\C{D}_B^i = \argmin_{\C{D}_B^i \subset P^i} \sum_{x \in P^i} d(\C{D}_B^i \cup \C{D}_L^i, x)$\\
    Update all sets and train the classifier:\\
      \advance\leftskip by 1ex $\C{D}_L^{i+1} \leftarrow \C{D}_L^i \cup \C{D}_B^i$ \hspace{1cm}
      $\C{D}_U^{i+1} \leftarrow \C{D}_U^i \setminus \C{D}_B^i$ \hspace{1cm}
      $h^{i+1} \leftarrow h^i + \C{D}_B^i$
  }
  \label{alg:iwkmeans}
  \caption{IWKmeans algorithm}
 \end{algorithm}

\subsubsection{Potential concerns.} The fact that the method repels all selected samples and not only the noisy ones can be debated.
We tested variants of this method that repels noisy samples only, or noisy and very easy to classify samples as they can also be considered detrimental \cite{abraham2020rebuilding}.
Since all variants had similar performances, we present here the simplest one.
Another concern is the convergence of this modified version of K-Means. It is easy to imagine in two dimensions how \emph{fixed} centers can prevent a \emph{moving} one to reach its minimum. From our experience, the K-Means++ initialization prevents most of these problems, and Fig.~\ref{fig:synthetic} proves the method's efficiency in a two-dimensional setting. For the sake of clarity and concision, we refer the reader to this online study of IWKMeans convergence\footnote{\url{https://dataiku-research.github.io/cardinal/auto_examples/plot_incr_kmeans.html}}.

\section{Experiments}
\label{sec:exp}

We perform active learning experiments on synthetic and natural datasets following the framework described in \cite{abraham2020rebuilding}. \emph{Random sampling} (Random) is the baseline. We use \textit{KCenterGreedy} (KCenter) as a proxy for Core-sets \cite{sener2017active} since there is no open implementation available. Note that the latter uses the activation of the penultimate layer of neural networks, so we have adapted it to random forests by considering a PCA-reduced forest embedding. We compare \emph{lowest confidence sampling} (Confidence) as described above to its informed counterpart \textit{IConfidence}. We also compare \emph{Weighted K-Means}\cite{zhdanov2019diverse} (WKMeans) with $\beta=10$ to our proposed \textit{IWKMeans}. BatchBALD\cite{kirsch2019batchbald} was not considered due to its prohibitive computational time of several hours compared to less than one minute for others.

We run ten iterations using five repeated two-fold cross-validation for each task. Reported results include means and confidence intervals at 10th and 90th quantiles. Cifar10 and Cifar100 tasks are run on ImageNet embeddings, Cifar10 SimCLR is run on embeddings learned using contrastive learning \cite{chen2020simple}, and other tasks are run using raw data. A Random Forest is used on the LDPA task, all others use a multi-layer perceptron with hidden layers of size 128 and 64. More details can be found on the code repository\footnote{\url{https://github.com/dataiku-research/sample_noise_impact_on_active_learning}} or in \cite{abraham2020rebuilding}.

\subsection{Synthetic problem with noisy samples}

\begin{figure}[htp]
  \centering
  \subfloat[2 features\label{fig:subim1}]{%
    \includegraphics[width=0.495\textwidth]{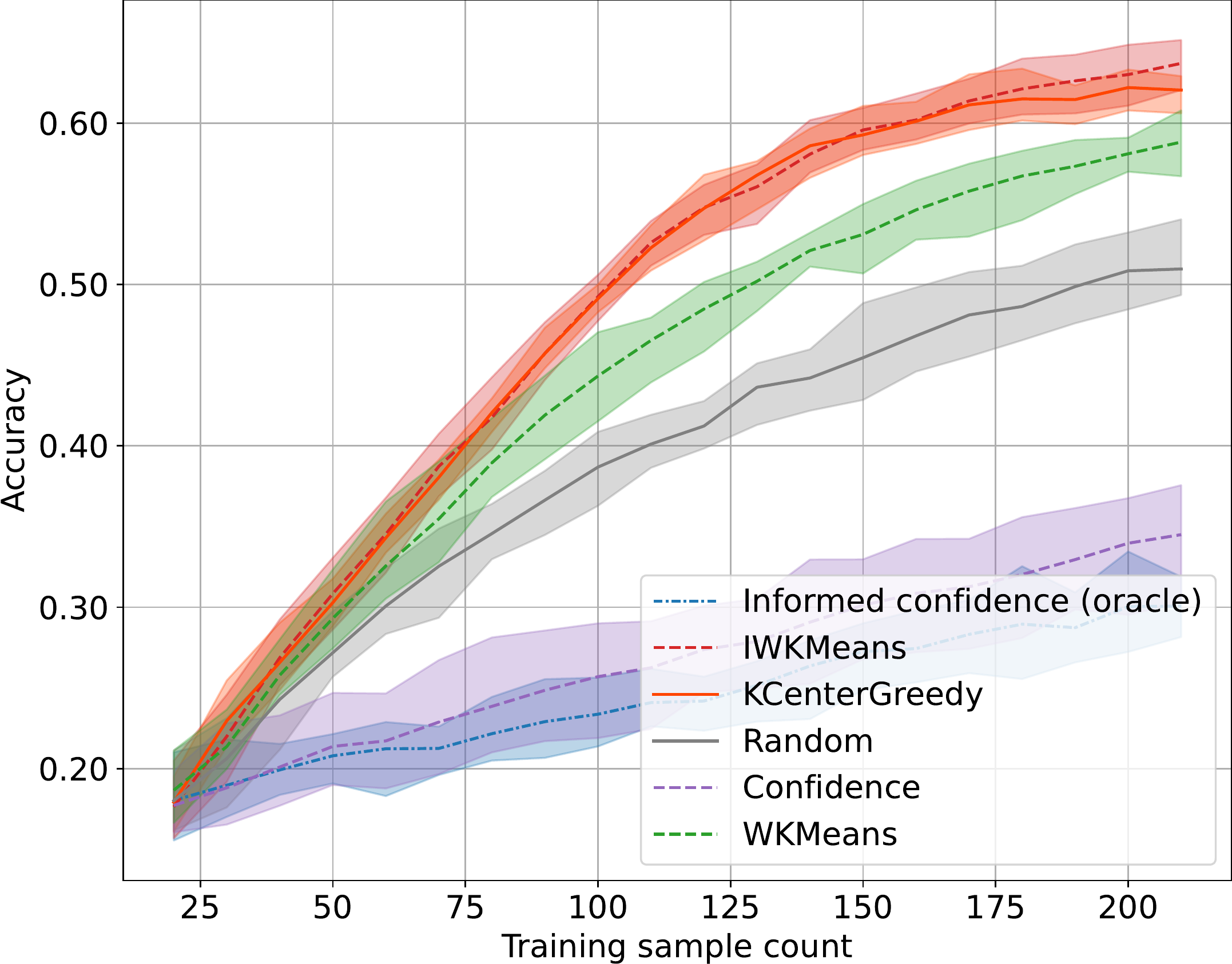}%
  }\hfill
  \subfloat[40 features\label{fig:subim2}]{%
    \includegraphics[width=0.495\textwidth]{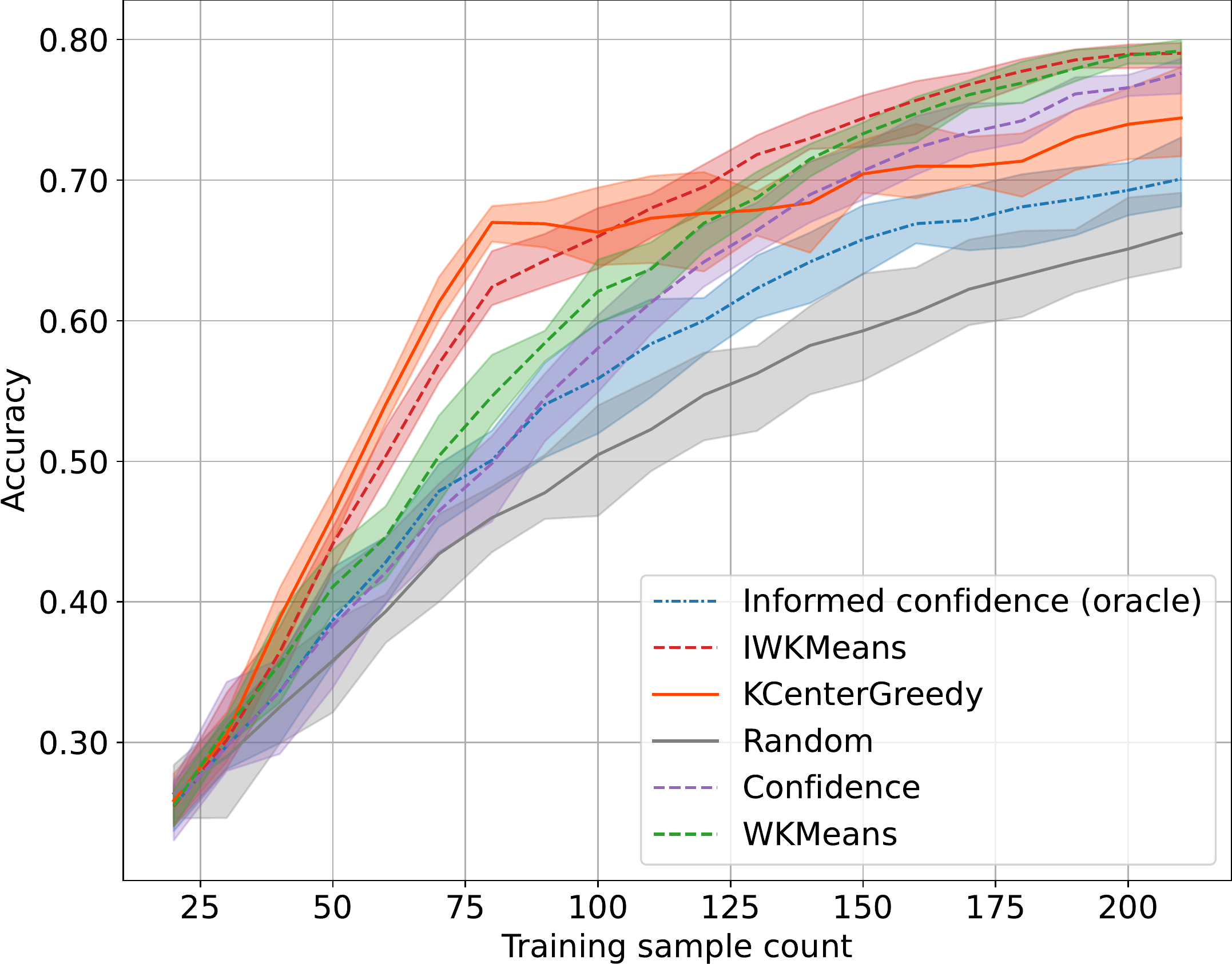}%
  }
  %\subfloat[40 features\label{fig:subim2}]{%
  %  \includegraphics[width=0.5\textwidth]{images/noisy_noisy_in_selected_boxplot.pdf}%
  %}
  
  \caption{Test accuracy on synthetic problems.}
  \label{fig:synthetic}
  
\end{figure}

To create noisy samples, we design a task where samples from a given class are not distinguishable
from those of another class. We create a 10-class task composed of spatially isolated blobs. 
Some blobs are composed of regular samples that all belong to the same class.
Other blobs are composed of samples randomly assigned to two different classes; we call them
noisy blobs since their samples are impossible to classify.
We create a low-dimensional problem with 10000 samples, 2 features, 10 classes, 200 blobs,
half of which are noisy. The active learning experiment uses 20 batches of 20 samples. We also
create a high-dimensional problem with the same characteristics except that the data has 40 features, and
we generate 90 blobs, 30 of which are noisy. We use accuracy AUC over the whole experiment to measure strategy performances.
In this synthetic experiment, we know which samples are noisy by construction and therefore report the ratio of noisy samples (NSR) as a measure of sample noise instead of its proxy RBA. Note that RBA is strongly correlated ($> 0.95$) with NSR.
Results are reported in Fig.~\ref{fig:synthetic}.

\begin{table}[htp]

  \caption{AUC and ratio of noisy samples per method. Standard deviation is in parenthesis. Best answers in terms of accuracy (higher) and Noisy Sample Ratio (lower) are in bold.}
  \label{tab:synthetic}
\centering
\begin{tabular}{llcccccc}
  \hline
  Dataset  & Metric                & Random        & KCenter          & Confidence      & IConfidence     & WKMeans           & IWKMeans            \\
  \hline
  Noisy LD & AUC                   &    38.6 (1.5) &       47.9 (0.5) &    26.7 (2.5)   &    24.5 (1.4)   &        44.0 (1.2) & {\bf 48.1} (1.0)    \\
  Noisy LD & NSR                   &    50.3 (4.1) &       42.4 (2.0) &    38.9 (6.5)   &{\bf 10.1} (4.6) &        43.5 (2.6) &      39.3  (1.8)    \\
  \hline
  Noisy HD & AUC                   &    50.7 (2.1) &       61.7 (1.1) &    58.0 (1.2)   &    55.0 (1.5)   &        60.6 (0.9) & {\bf 63.2} (0.6)    \\
  Noisy HD & NSR                   &    35.0 (3.0) &       24.5 (1.5) &    25.6 (1.5)   &{\bf 3.2} (1.1)  &        33.4 (1.5) &      26.9  (1.8)    \\
  \hline
  \end{tabular}
\end{table}
  
In terms of performances, IWKMeans dominates all methods, which is what was expected. KCenter is closely
following which is surprising since the model here is a random forest and we did not expect our quick
adaptation to this model to perform well.
We would have expected Confidence to select more noisy samples and perform poorly because of that. Instead, it seems to be penalized by its lack of diversity and exploration.
IConfidence minimizes the number of noisy samples selected, as expected, and yet it performs as badly as Confidence for the same reasons. In the end, this experiment shows that diversity can be as crucial as sample noise, and we expect a sweet spot to exist. Overall, we also observe that IWKMeans seem to be more robust to noisy samples. More insights are available in appendix Fig. \ref{fig:synthetic_noisy}.

\begin{figure}[hbtp]
  \centering
  \subfloat[CIFAR 10, ImageNet embedding\label{fig:cifar10}]{%
    \includegraphics[width=0.495\textwidth]{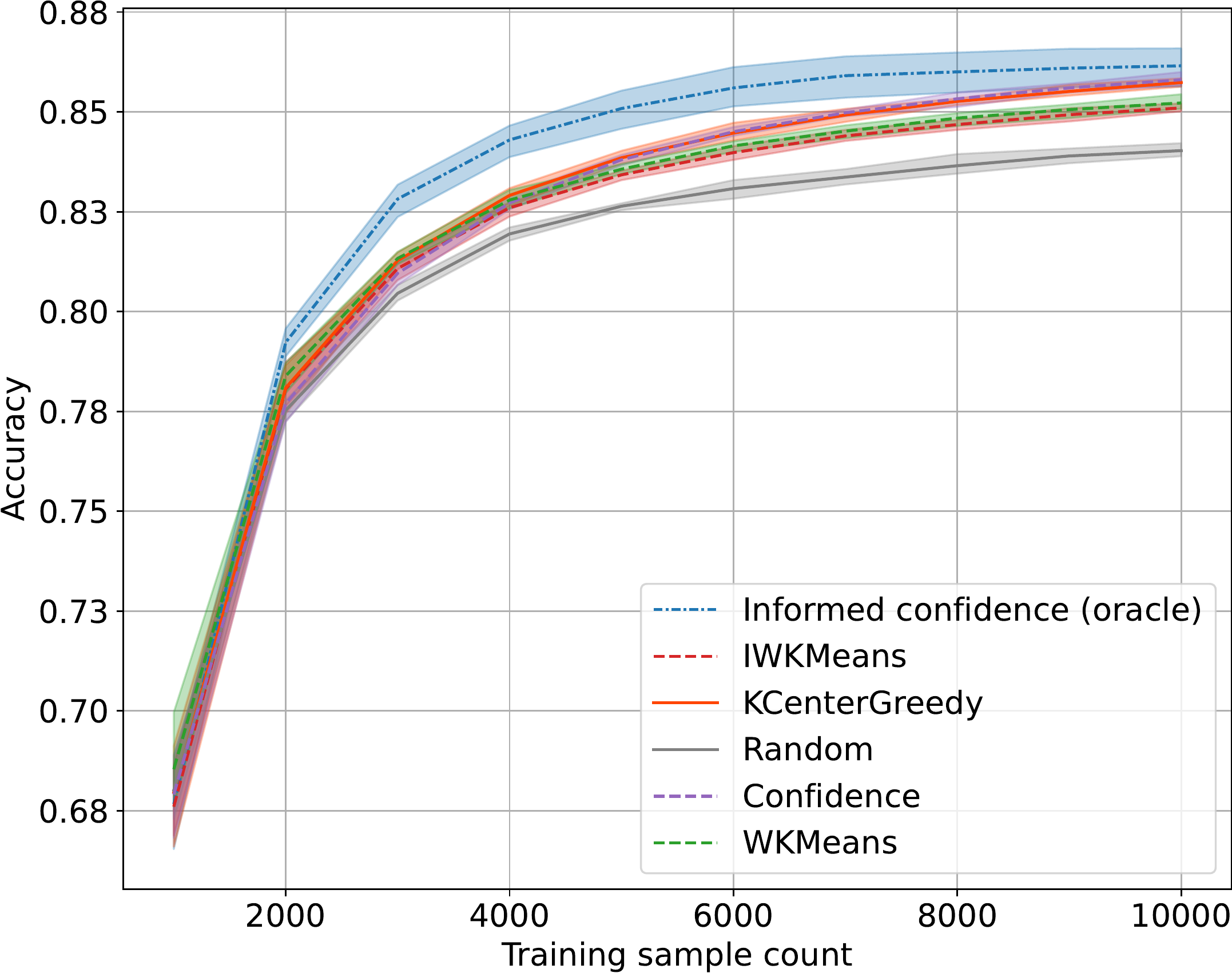}%
  }
  \hfill
  \subfloat[CIFAR 10, SimCLR embedding\label{fig:cifar10_sim}]{%
    \includegraphics[width=0.495\textwidth]{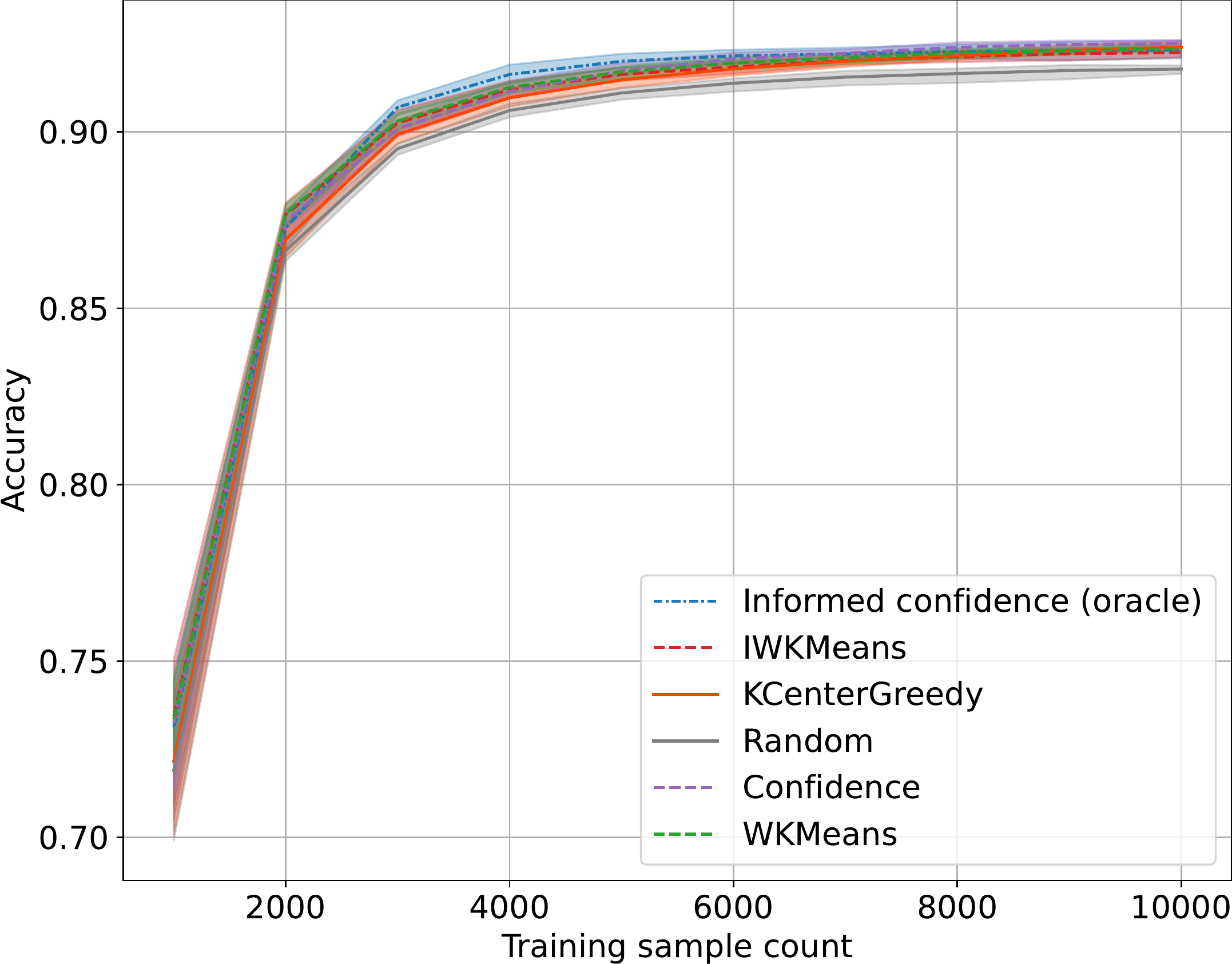}%
  }\\
  \subfloat[CIFAR 100\label{fig:cifar100}]{%
    \includegraphics[width=0.495\textwidth]{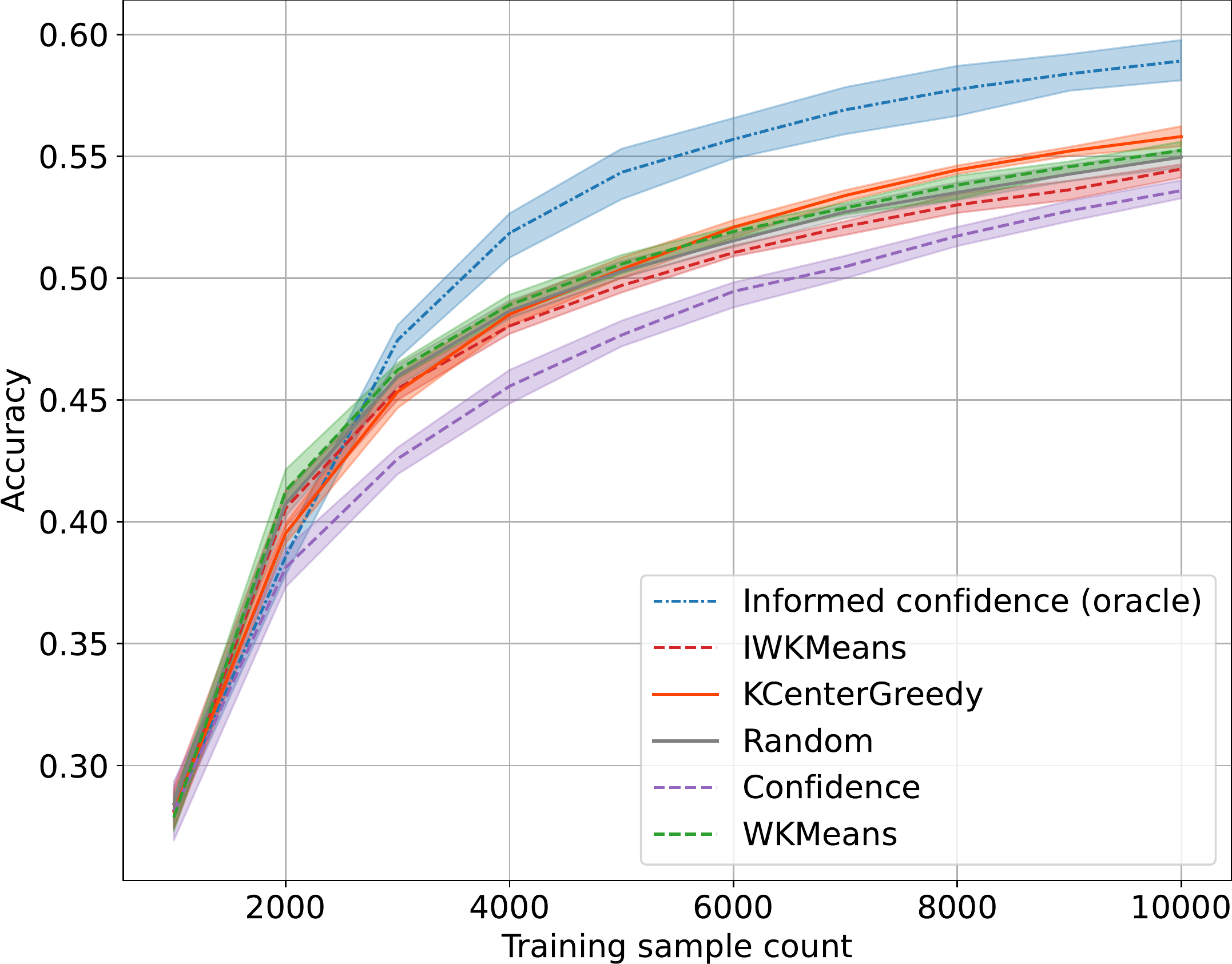}%
  }\hfill
  \subfloat[LDPA\label{fig:ldpa}]{%
    \includegraphics[width=0.495\textwidth]{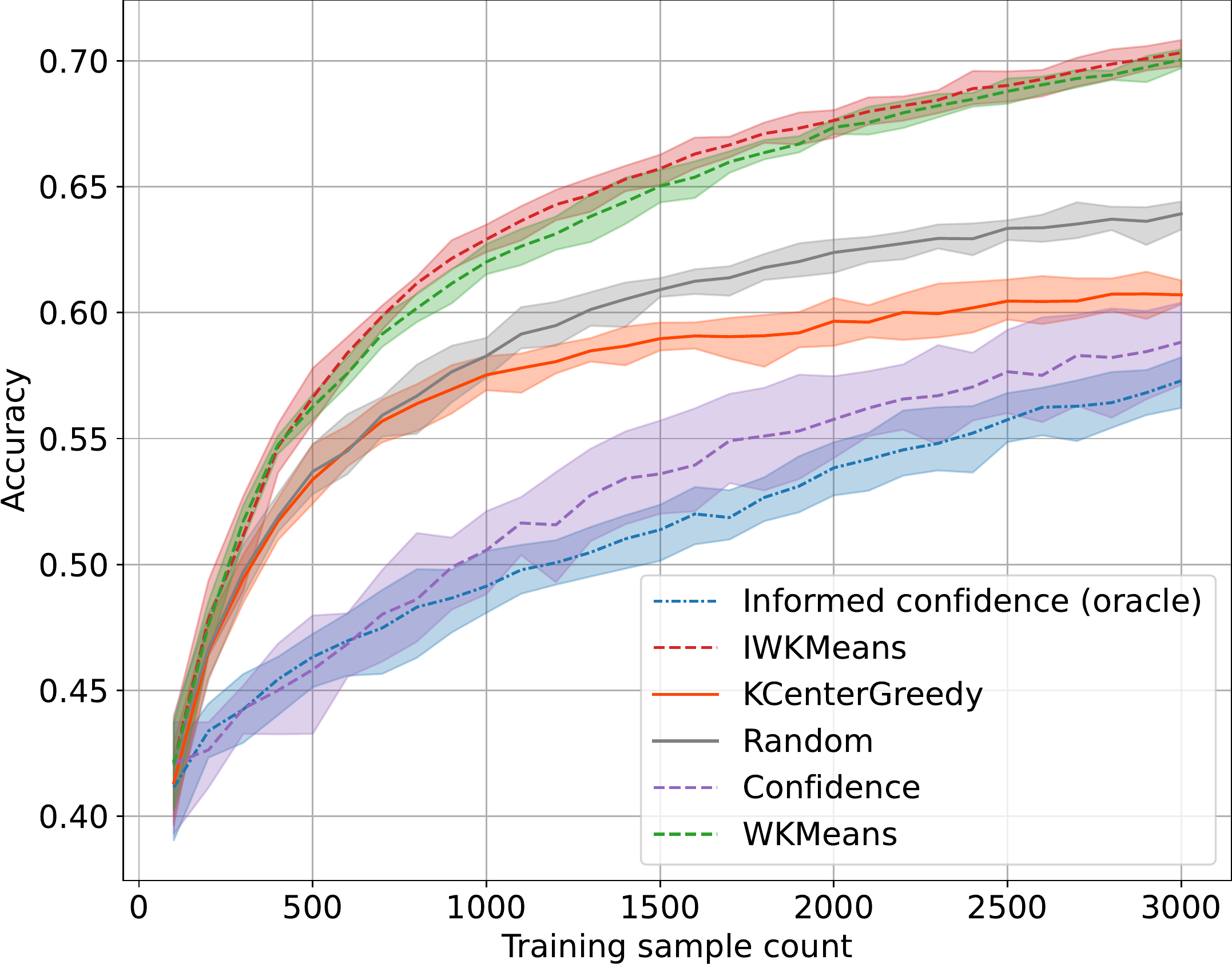}%
  }\\
  \subfloat[MNIST\label{fig:mnist}]{%
    \includegraphics[width=0.495\textwidth]{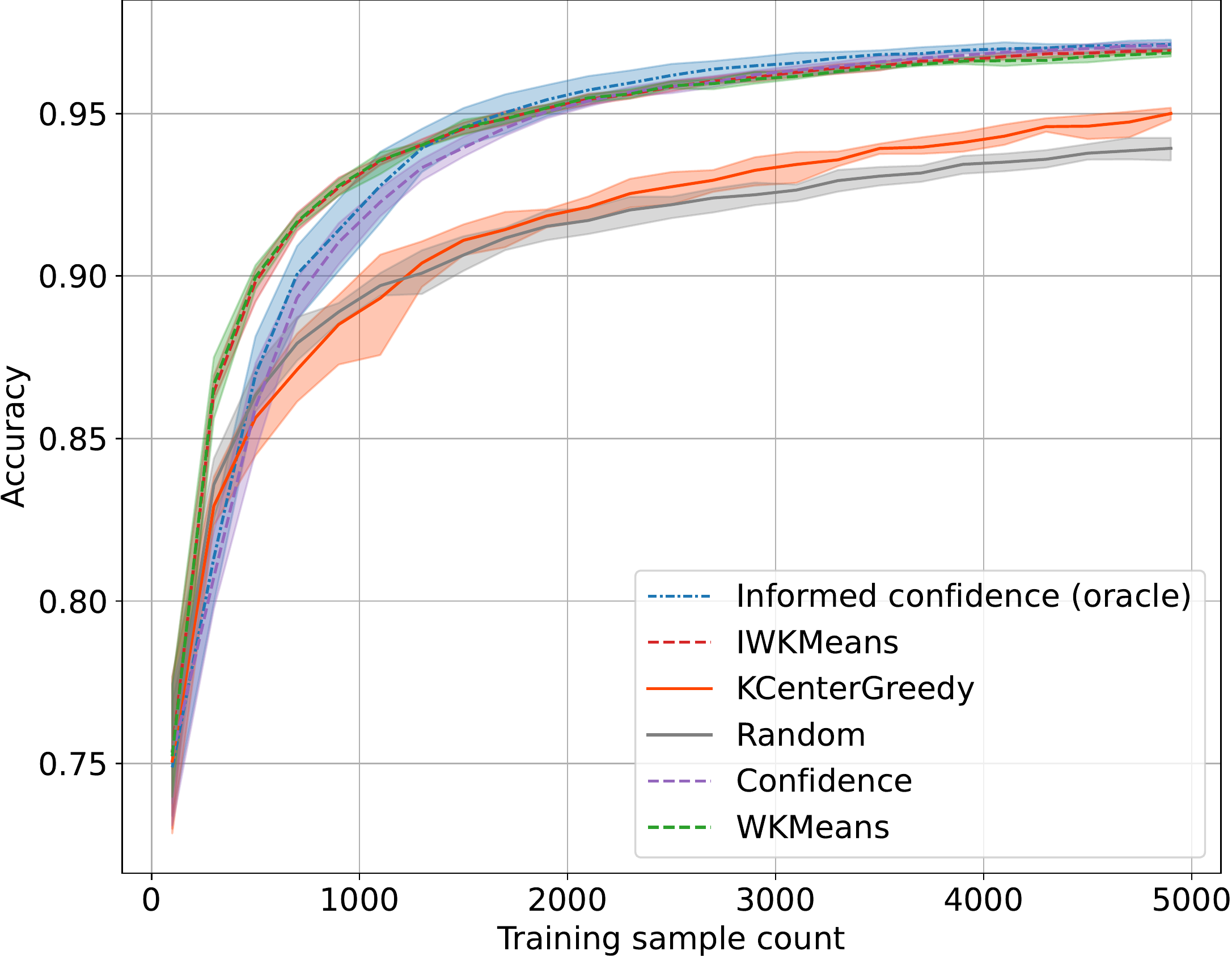}%
  }\hfill
  \subfloat[Fashion MNIST\label{fig:fashion}]{%
    \includegraphics[width=0.495\textwidth]{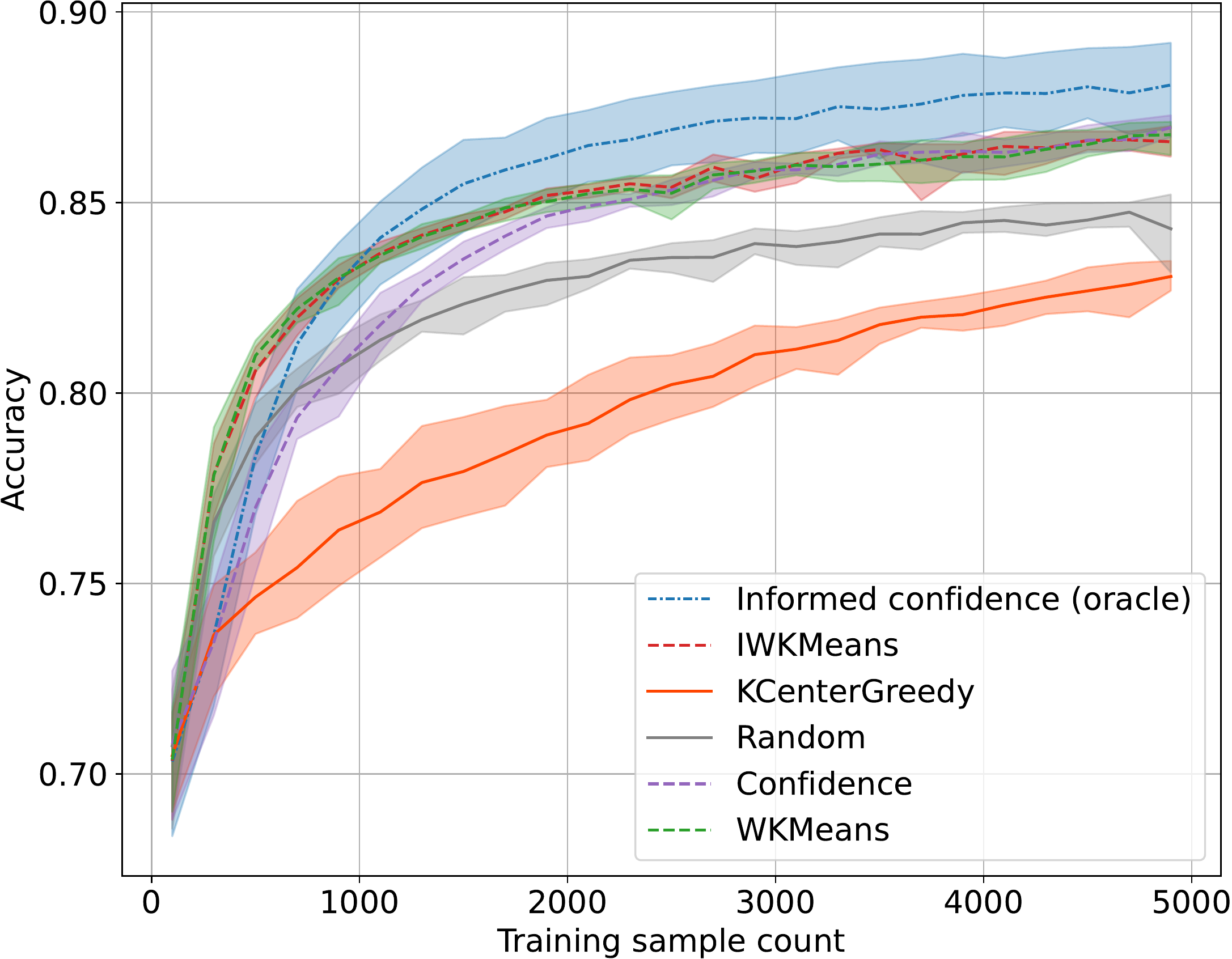}%
  }

  \caption{Results on real datasets}
  \label{fig:image2}
  
\end{figure}

\begin{table}[hbtp]

  \caption{Area under the curve for accuracy (AUC) and reverse batch accuracy (RBA) per method averaged over all repetitions. Standard deviation is in parenthesis. Bold values are statistically significantly higher than the others based on a Friedman test with Nemenyi post-hoc test which details are available in Fig. \ref{fig:significance} in appendix.}
  \label{tab:synthetic}

  \centering

\begin{tabular}{llcccccc}
  \hline
  Dataset  & Metric                & Random          & KCenter          & Confidence       & IConfidence     & WKMeans         & IWKMeans         \\
  \hline          
  LDPA     & AUC                   & {\bf 59.0} (0.5)&    57.2 (0.5)    &   51.9  (1.1)    &    51.2  (0.8)  & {\bf 63.1} (0.3)& {\bf 63.6} (0.3) \\
  LDPA     & RBA                   &    67.1 (0.7)   &    49.3 (2.3)    &   51.6  (2.0)    &{\bf 98.9} (0.1) &    67.8 (1.1)   &       67.6 (1.1) \\
  \hline          
  Cifar10  & AUC                   &    80.9 (0.2)   & {\bf 82.0} (0.2) & {\bf 81.9} (0.2) &{\bf 82.9} (0.4) &    81.8 (0.2)   &       81.6 (0.2) \\
  Cifar10  & RBA                   &    91.5 (4.8)   &   81.5 (10.7)    &   80.5 (12.6)    &{\bf 94.9} (3.5) &    85.2 (9.0)   &       85.3 (9.1) \\
  \hline   
  Cifar10S & AUC                   &    88.8 (0.2)   &   89.2  (0.2)    & {\bf 89.5} (0.2) &{\bf 89.6} (0.3) &{\bf 89.4} (0.2) & {\bf 89.5} (0.3) \\
  Cifar10S & RBA                   &    93.5 (1.3)   &   87.5  (1.8)    &   80.0  (3.6)    &{\bf 96.5} (0.8) &    86.2 (2.8)   &       87.9 (2.3) \\
  \hline            
  MNIST    & AUC                   &    90.9 (0.2)   &   91.2  (0.3)    & {\bf 93.5} (0.2) &{\bf 93.8} (0.3) & {\bf 94.2} (0.1)& {\bf 94.2} (0.1) \\
  MNIST    & RBA                   &{\bf 97.6} (0.2) &   96.6  (0.4)    &   92.3  (8.1)    &{\bf 97.7} (2.5) &    88.1 (0.4)   &       86.9 (0.6) \\
  \hline            
  Fashion  & AUC                   &    82.4 (0.2)   &   79.3  (0.3)    & {\bf 83.5}  (0.3)&{\bf 85.0} (1.0) & {\bf 84.3} (0.1)& {\bf 84.3} (0.1) \\
  Fashion  & RBA                   &    88.1 (0.4)   &{\bf 90.8}  (9.7) &   82.3 (15.9)    &{\bf 91.3} (7.3) &    70.6 (0.7)   &       69.2 (0.7) \\
  \hline              
  Cifar100 & AUC                   &    48.5 (0.3)   & {\bf 48.3} (0.2) &   46.2  (0.2)    &{\bf 50.8} (0.6) & {\bf 48.9} (0.2)&       49.0 (0.3) \\
  Cifar100 & RBA                   &    69.4 (9.2)   &   71.2 (14.1)    &   55.6 (15.6)    &{\bf 88.8} (5.8) &    70.7 (9.2)   &       70.0 (9.9) \\
  \hline
\end{tabular}
\end{table}

\subsection{Real datasets}

We now analyze the samplers behaviors on our collection of real-life datasets.

\textbf{Informed lowest confidence.}
IConfidence is equivalent or better than confidence in all cases. It is also the best strategy for all tasks except MNIST and LDPA. Note that
the RBA of this method is much higher than the other strategies. It reveals that getting \emph{too close} to the decision boundary may not be required for good performance. Even more, this oracle method does not enforce diversity but yet overpowers diversity enforcing methods. This questions the fundamental hypothesis that enforcing diversity is mandatory to obtain good performances.
Further work will investigate further this sampler and try to reproduce its behavior online with proxy metrics proposed in \cite{abraham2020rebuilding}.

\textbf{IWKMeans.} WKMeans and IWKMeans bring a significant uplift against random and all other uncertainty-based or unsupervised methods in all tasks except
CIFAR10 with SimCLR embeddings. IWKMeans outperforms WKMeans on LDPA only, making it hard to draw a definitive conclusion on real tasks.
Further experiments are needed to investigate these behaviors. Early investigations suggest that the variation in density of noisy samples in
multiclass settings can tamper with adverse-to-noise samplers. For example, a general strategy can be hard to find on the MNIST dataset where few noisy samples
exist between classes zero and four, while their density is high between classes three and five.

\textbf{SimCLR embedding.} An unexpected conclusion of these experiments is that contrastive-based embeddings can bring an uplift significantly higher than choosing the best query sampling strategy.

\section{Conclusion}

In active learning, noisy samples that are hard to classify by the model can be detrimental to the performance.
To prove this, we have designed a metric to measure them and a synthetic problem to test the robustness of query strategies to their presence.
IWKmeans, the proposed noise-adverse sampling strategy, has been proven effective on synthetic data, but not on real tasks where it marginally improves WKMeans on which it is based. If IWKMeans' performance seems correlated to the number of noisy samples selected,
 there may be more than meets the eye in this problem, and more investigations are needed.
Our study also shows that a sampler as simple as confidence sampling can outperform all other samplers if
informed by a good enough classifier. Whether or not this uplift can be reproduced in real conditions using a
proxy must be investigated in further work.

\printbibliography

\clearpage
\appendix

\renewcommand{\thefigure}{A\arabic{figure}}

\section{Insights about sample noise in synthetic examples}

\begin{figure}[htp]
  \centering
  \subfloat[2 features\label{fig:subim1}]{%
    \includegraphics[width=0.495\textwidth]{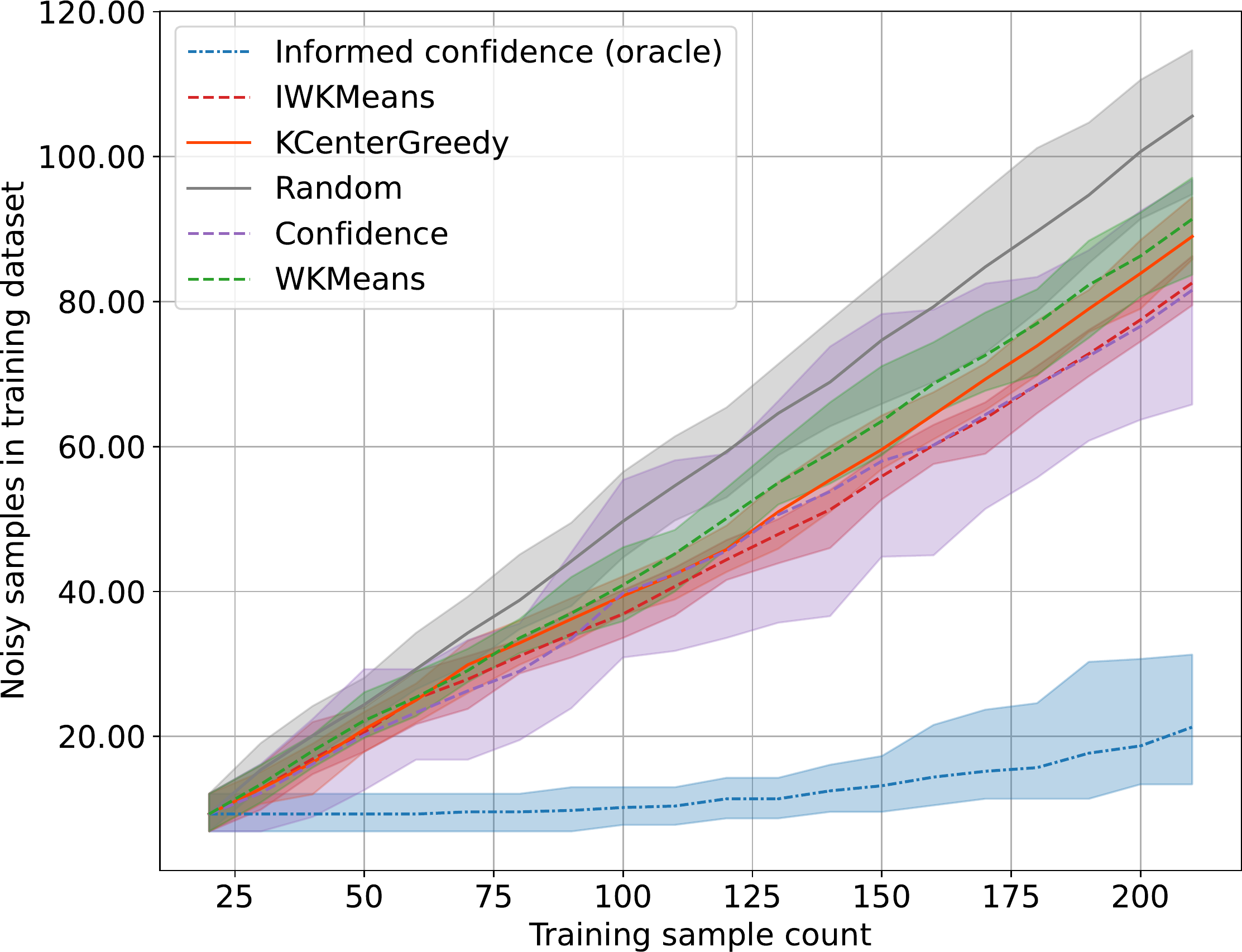}%
  }\hfill
  \subfloat[40 features\label{fig:subim2}]{%
    \includegraphics[width=0.495\textwidth]{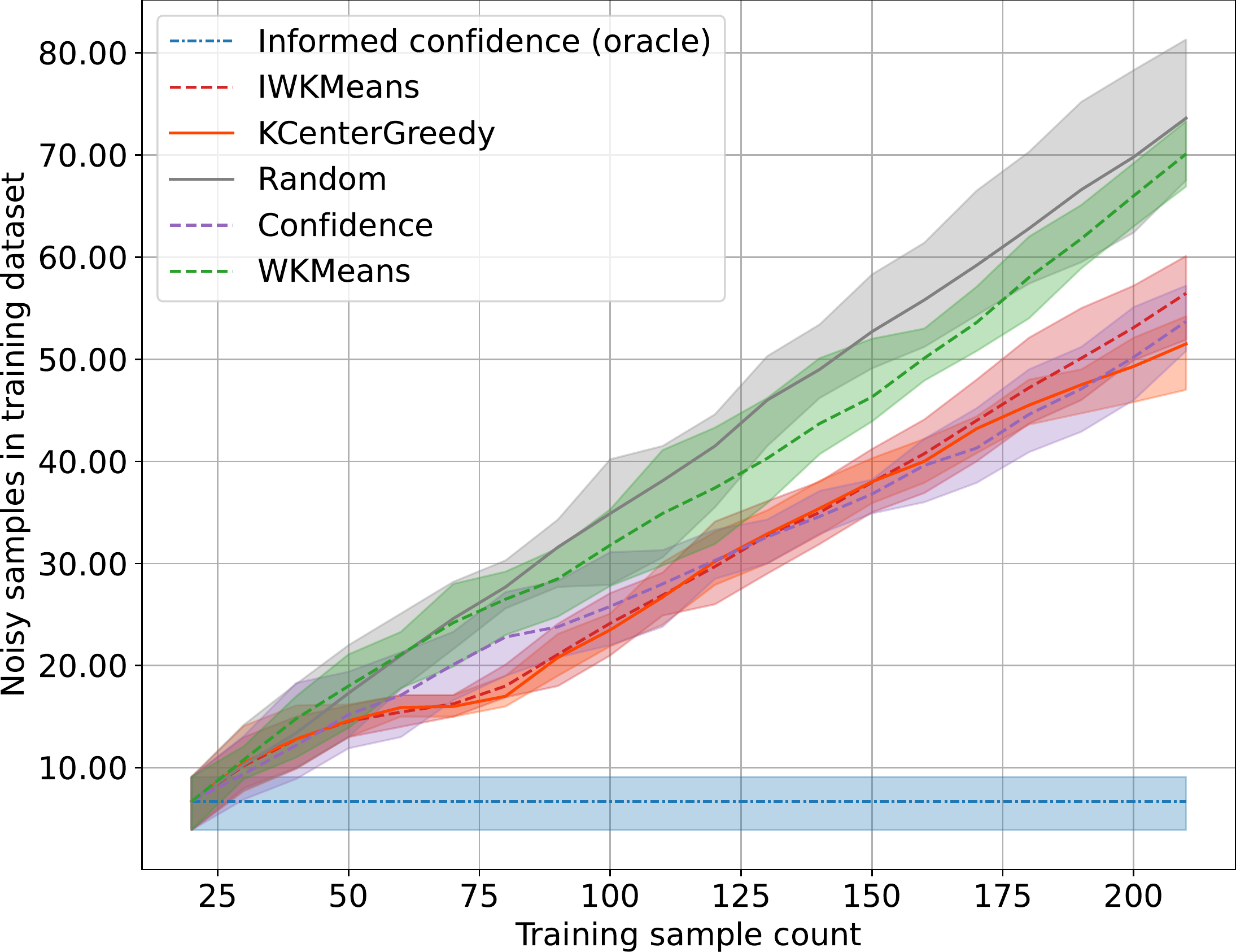}%
  }
  \caption{Number of noisy samples across experiments on synthetic tasks.}
  \label{fig:synthetic_noisy}
  
\end{figure}

During the whole active learning experiment, all methods keep a constant pace in the ratio of noisy samples in their training set, as shown in Fig. \ref{fig:synthetic_noisy}.

\newpage
\section{Statistical significance for each method on each task}

In order to determine the best methods for ech task, we rank the scores using the python package autorank. It performs a Freidman test followed by a Nemenyi post-hoc test.

\begin{figure}[hbtp]
  \centering
  \subfloat[CIFAR 10, ImageNet embedding\label{fig:cifar10_st}]{%
    \includegraphics[width=0.495\textwidth]{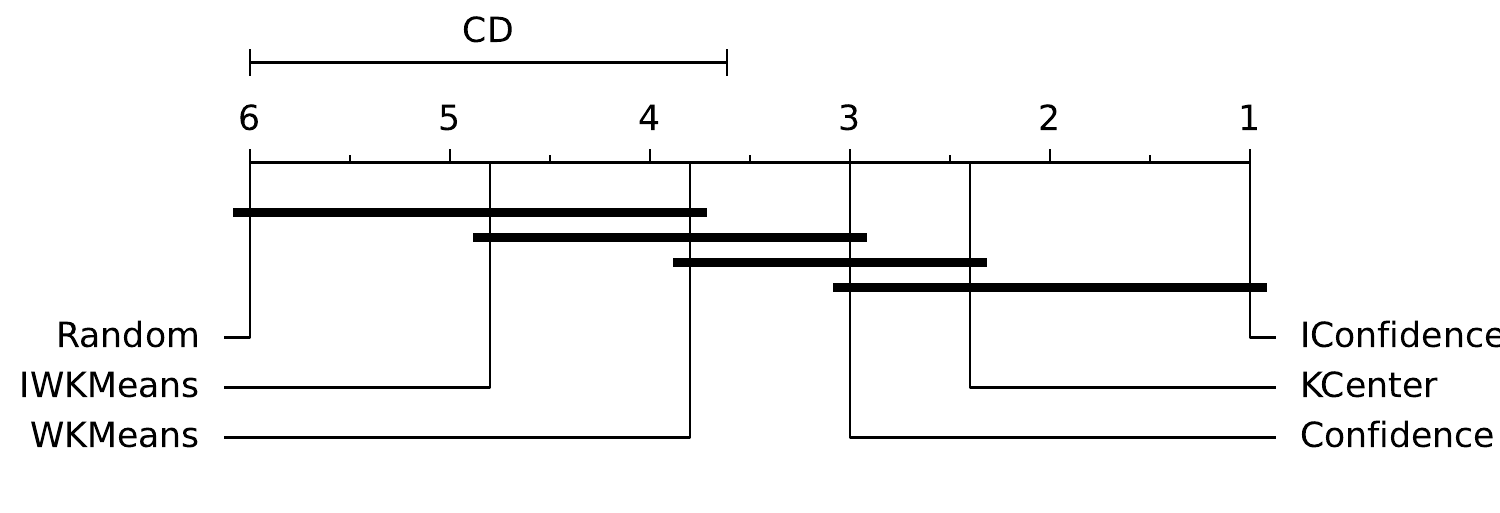}%
  }
  \hfill
  \subfloat[CIFAR 10, SimCLR embedding\label{fig:cifar10_sim_st}]{%
    \includegraphics[width=0.495\textwidth]{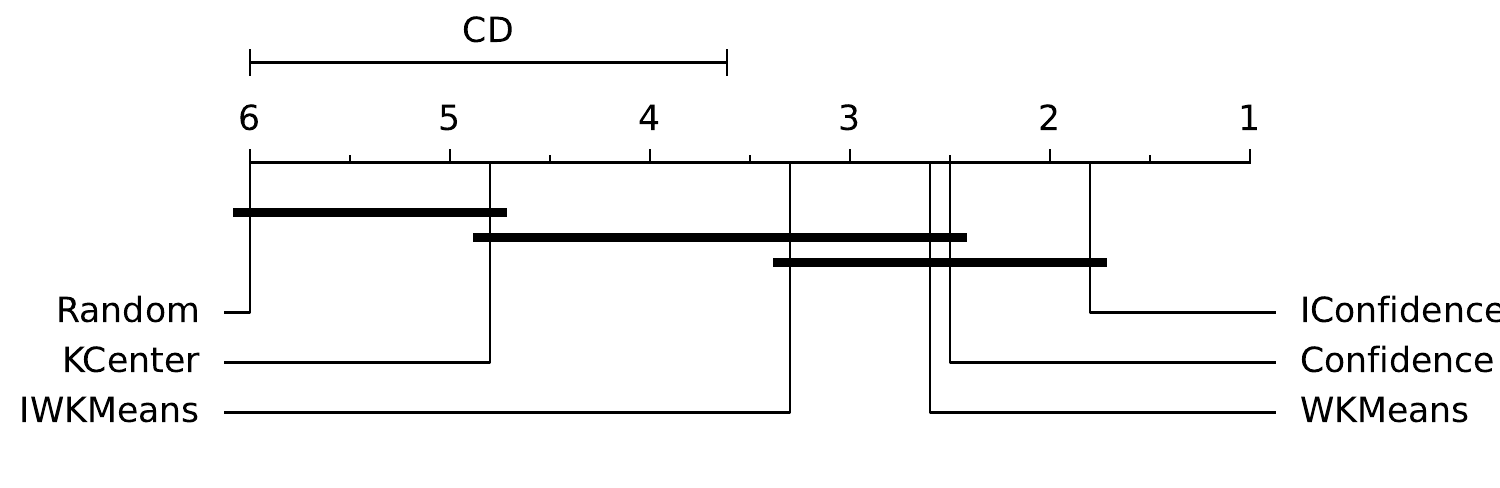}%
  }\\
  \subfloat[CIFAR 100\label{fig:cifar100_st}]{%
    \includegraphics[width=0.495\textwidth]{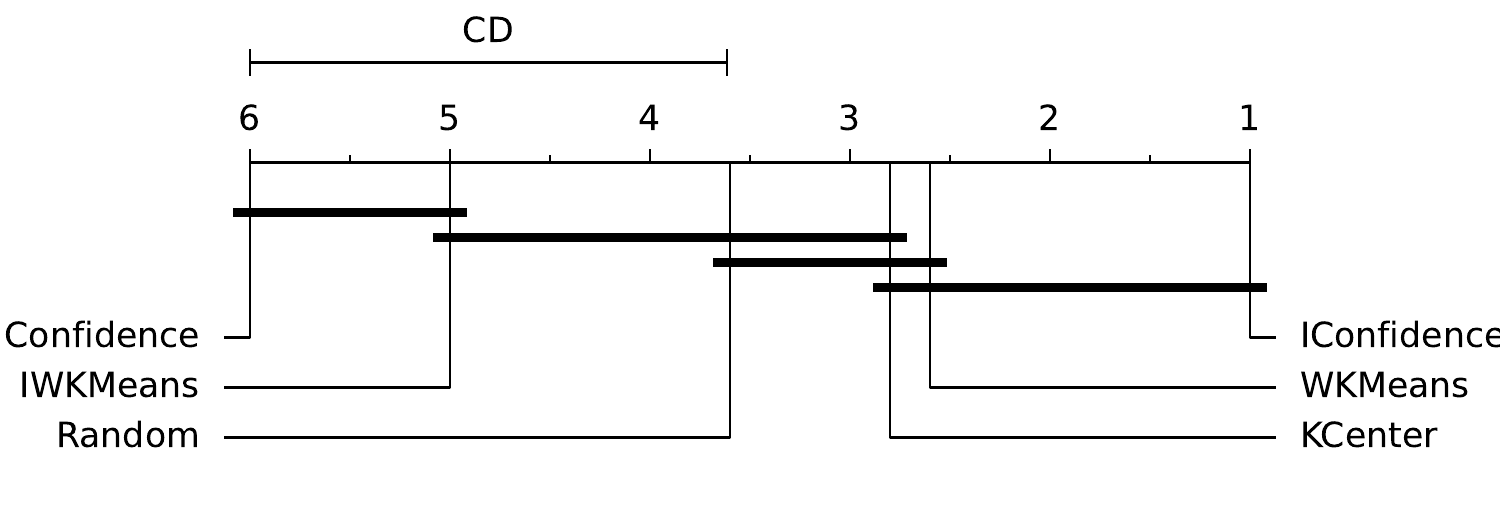}%
  }\hfill
  \subfloat[LDPA\label{fig:ldpa_st}]{%
    \includegraphics[width=0.495\textwidth]{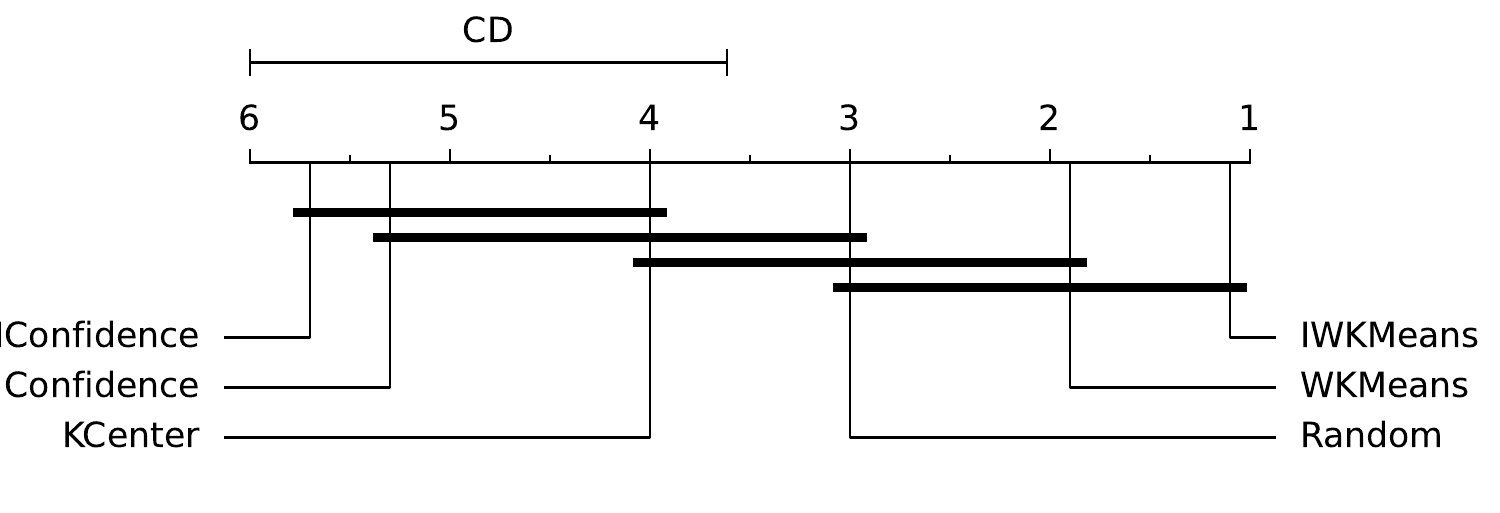}%
  }\\
  \subfloat[MNIST\label{fig:mnist_st}]{%
    \includegraphics[width=0.495\textwidth]{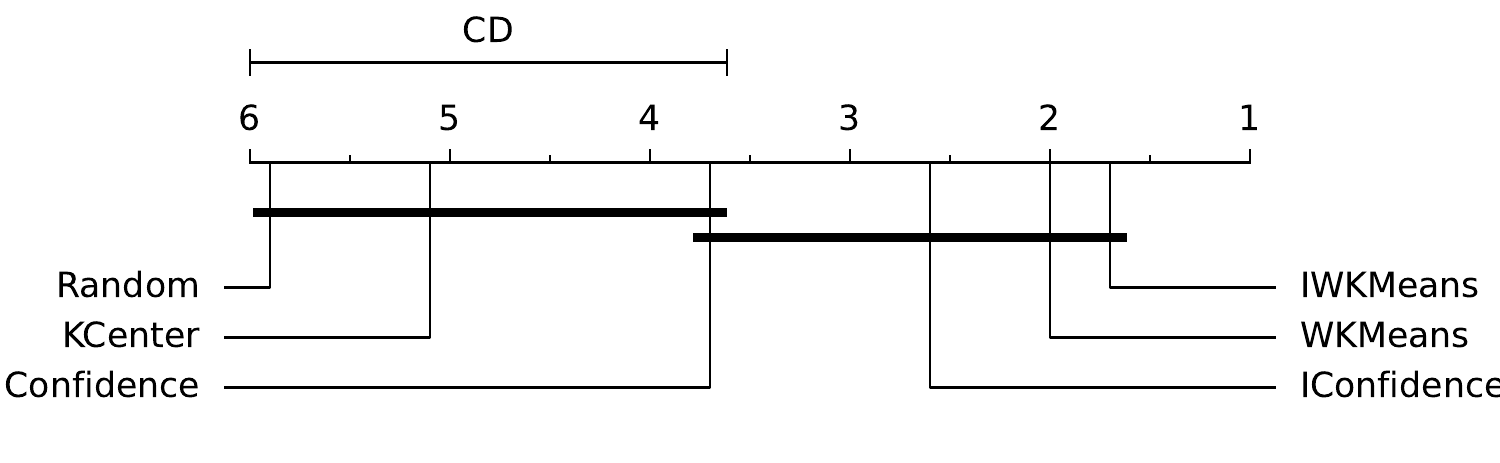}%
  }\hfill
  \subfloat[Fashion MNIST\label{fig:fashion_st}]{%
    \includegraphics[width=0.495\textwidth]{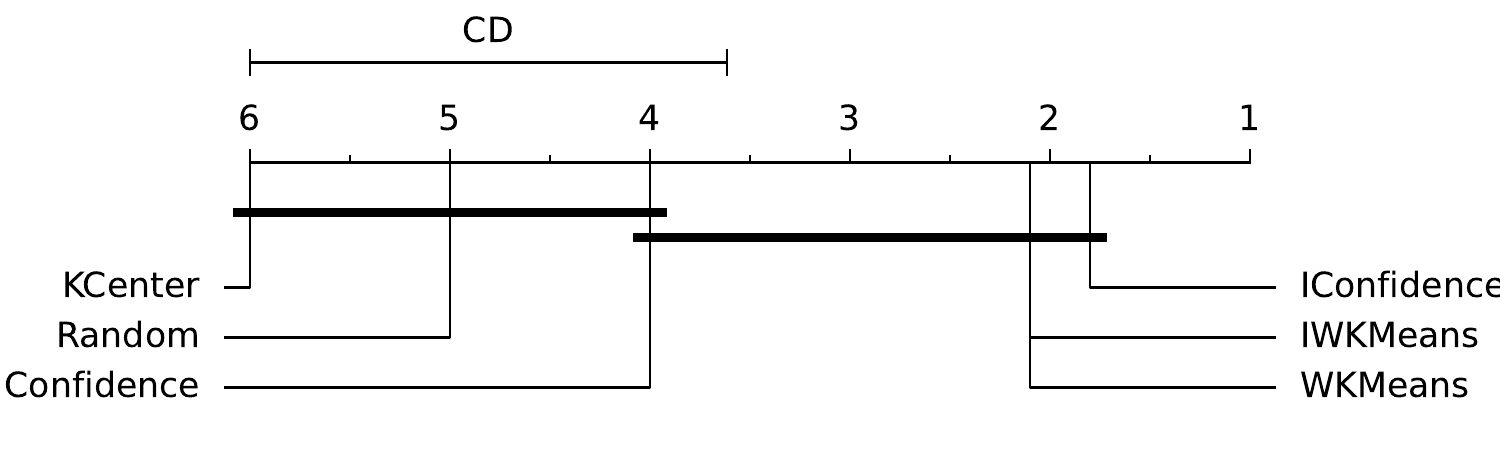}%
  }

  \caption{Results on real datasets}
  \label{fig:significance}
  
\end{figure}

\end{document}